\documentclass{article}

\PassOptionsToPackage{numbers, compress}{natbib}

\usepackage[final]{neurips_2023}

\usepackage[utf8]{inputenc} 
\usepackage[T1]{fontenc}    

\usepackage{url}            
\usepackage{booktabs}       
\usepackage{amsfonts}       
\usepackage{nicefrac}       
\usepackage{microtype}      
\usepackage{xcolor}         
\usepackage{algorithm}
\usepackage{algpseudocode}
\usepackage{amsmath}
\usepackage[pdftex]{graphicx}
\usepackage{wrapfig}

\title{Automatic Discovery of Visual Circuits}

\definecolor{myBlue}{RGB}{76,81,223}
\definecolor{niceRed}{RGB}{210,48,38}
\usepackage[colorlinks, citecolor=myBlue, linkcolor=niceRed]{hyperref}

\author{%
Achyuta Rajaram$^{1,2}$\thanks{Indicates equal contribution. Correspondence to achyuta@mit.edu, nchow@mit.edu, schwett@mit.edu.} \quad Neil Chowdhury$^{2}$\footnotemark[1] \\ 
\textbf{Antonio Torralba}$^2$ 
\quad \textbf{Jacob Andreas}$^2$ 
\quad \textbf{Sarah Schwettmann}$^2$ 
\\ 
$^1$Phillips Exeter Academy \quad $^2$MIT CSAIL\\
}

\begin{document}

\maketitle

\begin{abstract}

    To date, most discoveries of network subcomponents that implement human-interpretable computations in deep vision models have involved close study of single units and large amounts of human labor. We explore scalable methods for extracting the subgraph of a vision model’s computational graph that underlies recognition of a specific visual concept. We introduce a new method for identifying these subgraphs: specifying a visual concept using a few examples, and then tracing the interdependence of neuron activations across layers, or their \textit{functional connectivity}. We find that our approach extracts circuits that causally affect model output, and that editing these circuits can defend large pretrained models from adversarial attacks. Our code and data are available at \url{https://github.com/multimodal-interpretability/visual-circuits}.

\end{abstract}

\section{Introduction}
\label{sec:introduction}
Deep neural networks extract features layer by layer, until these features lead to a prediction. In vision models, studying these features at the level of individual neurons has revealed a range of human-interpretable functions that increase in complexity in deeper layers: Gabor filters \cite{erhan2009visualizing} in the earliest convolutional layers are followed by curve detectors \cite{cammarata2020curve}, and later, units that activate for specific categories of objects \cite{zhou2014object, zeiler2014visualizing, bau2017network, olah2017feature, bau2020units, hernandez2022natural}. However, there are many important questions that the study of individual neurons leaves unanswered---for instance, whether one feature is used to compute another, or whether two features share a common backbone. A mechanistic understanding of these kinds of phenomena is useful for understanding how models make decisions, whether a model capability relies on some other capability, and attributing unwanted behavior to learned subcomputations. For example, learned spurious correlations between features and outputs leads to model failures such as misclassification of dermatological images containing rulers as malignant \cite{narla2018ruler}, classifying huskies as wolves due to the presence of snow \cite{ribeiro2016why}, and learning gender and racial biases from training data and applying them during inference \cite{manresayee2022assessing}. We want to be able to intervene on the computational subgraph underlying these types of model behaviors and edit the set of features a model is using to make a decision. How can we automatically detect circuits in vision models?

\paragraph {Circuit discovery in vision models.} Previously, circuits underlying the detection of specific concepts have been identified in vision models via manual aggregation of model weights \cite{olah2020zoom}. For example, Olah et al. \citep{olah2020zoom} uncover a car circuit in InceptionV1 \cite{szegedy2015going} by creating feature visualizations~\cite{olah2017feature} of individual units, pinpointing a car-detecting neuron in the \verb|mixed4c| layer, and finding that the three neurons in the previous \verb|mixed4b| layer with maximal weight magnitudes also represent car features: wheels, windows, and car bodies. While this technique can indeed recover specific algorithms encoded in the weights of trained networks, scaling circuit extraction to larger models and more complex tasks will require approaches that automatically identify both features of interest and relevant subgraphs. 

\paragraph {Automated circuit extraction.} Recent work investigating the subcomputations inside neural language models has focused on detection of circuits that execute specific functions, such as indirect object identification (IOI) and identifying numbers greater than an input token (Greater-Than) \cite{wang2022interpretability}. Approaches to automating circuit discovery include subnetwork probing, which learns a mask over model components using an objective that combines accuracy and sparsity \cite{cao2021low}, and Automatic Circuit DisCovery (ACDC), a pruning-based technique that removes edges from a computational subgraph based on their effect on the output distribution \cite{conmy2023automated}. Conmy et al. \cite{conmy2023automated} show that ACDC successfully recovers the same IOI circuit identified by human researchers in GPT2-Small. Motivated by the promise of automated approaches in language models, our work explores the extension of scalable circuit detection to the visual domain.

This paper introduces a method for automatically identifying circuits in vision networks based on \textit{functional connectivity} of neurons between layers, or the interdependence of their activations in response to a particular \textit{input} distribution. We define a circuit as a computational subgraph of a trained network that derives information from input features to construct an intermediate representation that later affects the output distribution. We are interested in intermediate modifications to input representations, instead  of subgraphs which are responsible for a majority of a given observed output behavior (e.g. the definition proposed in~\cite{wang2022interpretability}). This distinction allows for the direct targeting of circuits representing visual features that cannot be easily expressed as a function of the model output (e.g. selecting circuits via text in a CLIP-style image-text embedding model). Given this definition, we introduce a new algorithm based on Cross-Layer Attribution (CLA) that iteratively refines circuit subgraphs based on attribution scores calculated between units in successive layers (Section \ref{sec:methods}).

To evaluate intermediate concept representation in CLA circuits, we construct CatFish, a dataset of composite images designed to be recognizable by simple circuits that operate over high-level visual features.
Intervention experiments on an InceptionV1 model finetuned on CatFish find that CLA automatically discovers circuits corresponding to intermediate concepts, and recovers known compositional relationships from CatFish within the model (Section \ref{sec:experiments}). We additionally apply our method to defend CLIP from text-based adversarial attacks using circuit interventions (Section \ref{sec:clip}). Together, these experiments show that CLA provides a simple and general mechanism for identifying functional dependencies between learned features in deep networks trained for computer vision tasks.

\vspace{-2mm}
\section{Methods}
\label{sec:methods}

\subsection{Circuit extraction based on Cross-Layer Attribution (CLA)}
\label{sec:CLA}

\begin{figure}[t!]
    \centering
    \includegraphics[width=\linewidth]{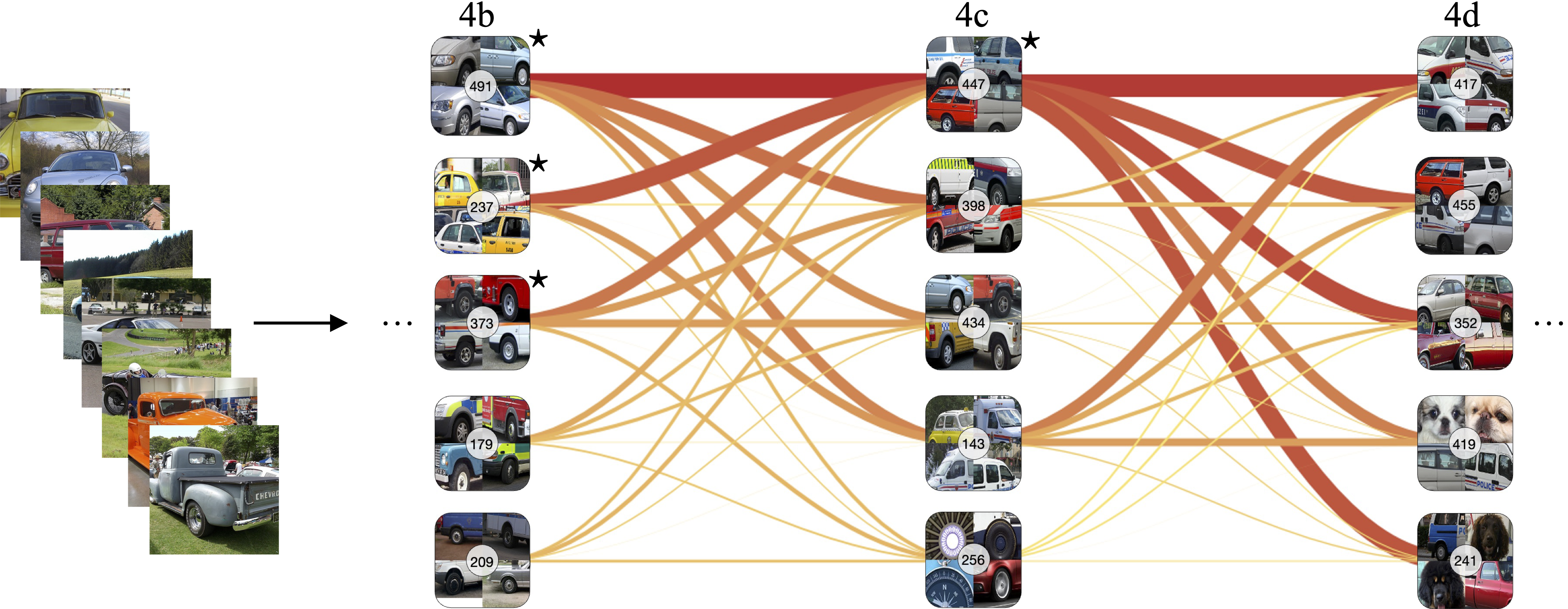}
    \vspace{-2mm}
    \caption{\textbf{Automatic discovery of the car circuit inside Inception using CLA.} The $\star$ indicates that a unit is present in the weight-based circuit discovered by Olah et al. in \cite{olah2020zoom}. Maximally activating dataset exemplars are shown for each neuron. CLA recovers all units in the car circuit (units 491, 237, 373 in Layer 4b; unit 447 in Layer 4c) from \cite{olah2020zoom}, as well as additional car-detecting neurons in all three layers studied. Edge thickness is proportional to Cross-Layer Attribution score.}  
    \label{fig:teaser}
    \vspace{-4mm}
\end{figure}
 
There are many possible ways to define a circuit. A purely structural notion of connectivity, for example, focuses on edges (weights) between learned features, such as in \cite{olah2020zoom}. To identify the circuit responsible for performing an arbitrary computation (e.g. detecting a concept) specified in model input, our approach instead focuses on a \emph{functional} notion of connectivity, based on which features influence the computation of other features in the input distribution.
Most existing attribution methods either explain how internal features are derived from inputs \cite{zhou2016learning, selvaraju2017grad}, or determine which features induce a distribution over outputs \cite{buildingblocks, schwettmann2023multimodal, conmy2023automated}.
We instead perform attribution \textit{between} internal layers, and iteratively refine a \textit{functional connectivity graph} by computing the set of features in a given layer (e.g. $l_i$) that maximally affect downstream features (in $l_{i+1}$). Experiments in Section \ref{sec:experiments} compare CLA to alternative circuit-discovery methods including the weight-based approach from \cite{olah2020zoom}.

\begin{algorithm}
\caption{\textbf{Cross-Layer Attribution (CLA)} computes a model subgraph corresponding to an input distribution in two steps. First, we compute an \textit{attribution matrix} which maps the functional connectivity between neurons in a set of layers. Then, we iteratively refine a candidate subgraph by updating the set of neurons included per layer to be those that maximize total attribution with neurons in the subgraph in the previous and following layers.}

\label{alg:cla}
\begin{algorithmic}[1]

\Procedure{AttributionMatrix}{model, layer $l_i$, layer $l_{i+1}$}
    \For{each input $x$}
        \State $a_i$ $\leftarrow$ model($x$).$l_i$ \Comment{Get $l_i$ activations on image $x$}
        \For{each neuron $m \in l_i,$ $n \in l_{i+1}$}
        \State attr$[x, m, n]$ $\leftarrow |a_{i, m}| \cdot \frac{\partial{\|l_{i+1, n}(a_i)\|_2}}{\partial{a_{i,m}}}$ \Comment{ Attribute $l_i$ neurons to $l_{i+1}$ neurons}
        \EndFor
    \EndFor
    \State~
    \Return ${\text{mean}}(\text{attr}, dim=0)$ \Comment{Compute mean attribution across images}

\EndProcedure
\\

\Procedure{BuildCircuit}{model, layers $\{l_i\}$, sizes $\{k_i\}$}
    \For{each $i$ in 1 to $L-1$}
        \State attrs[$l_i$] $\leftarrow$ \Call{AttributionMatrix}{model, $l_i$, $l_{i+1}$}
    \EndFor
    \State circuit[$l_1$] $\leftarrow$ (top $k_1$ argmax)$_n$ attrs$[l_{1},:,n]$    
\Comment{Initialize circuit with top neurons}
    \For{each $i$ in $2$ to $L$}
        \State circuit[$l_i$] $\leftarrow$ (top $k_i$ argmax)$_n$ sum(attrs$[l_{i-1},m_j,n]$, $m_j \in \text{circuit}[l_{i-1}]$)
    \EndFor
    \State prev circuit  $\leftarrow$ Ø
    \While{circuit $\neq$ prev circuit} \Comment{Iteratively refine circuit until it stops changing}
        \State prev circuit  $\leftarrow$ circuit
        \For{$i$ = $L-1$ to 1}
        \State circuit[$l_i$] $\leftarrow$ (top $k_i$ argmax)$_m$ sum(attrs$[l_i,m,n_j]$, $n_j \in \text{circuit}[l_{i+1}]$)
        \EndFor
        \For{$i$ = 2 to $L$}
            \State circuit[$l_i$] $\leftarrow$ (top $k_i$ argmax)$_n$ sum(attrs$[l_{i-1},m_j,n]$, $m_j \in \text{circuit}[l_{i-1}]$)
        \EndFor
    \EndWhile
    \State~
    \Return circuit
    
\EndProcedure

\end{algorithmic}
\end{algorithm}

Algorithm \ref{alg:cla} describes CLA. First, we compute attribution scores between neurons in subsequent layers. The score for a pair of neurons $(m\in l_i,$ $n\in l_{i+1})$ takes into account both relevance and influence, by multiplying together two terms: one corresponding to the magnitude of the activations, and one corresponding to the gradient of activations across layers. By computing attribution scores across all pairs of neurons, we create an input-distribution-dependent notion of cross-layer connectivity, termed an \textbf{attribution matrix}. Next, we build each circuit layer-by-layer using the attribution matrix, selecting a subset of $k$ neurons in each layer.  To do this, we first compute an ``intial guess'' of the first layer, naively choosing the top $k$ neurons in each layer, when ranked by the total sum of their attribution scores across the entire next layer, i.e. $\operatorname{arg\,max}_{n}^{(k)}\, \sum \text{attrs}[1, : , n]$. 
From there, we compute the entire circuit, selecting a subset of neurons in each layer which maximises the sum of attributions to the previous layer. We iteratively refine the resulting circuit by ``sweeping'' through the layers, re-selecting neurons to maximise the total sum of their attribution scores to the neurons \textit{within the circuit} in the successive and previous layers.

We show that CLA automatically recovers the ``car detector'' circuit previously identified manually in InceptionV1 \cite{olah2020zoom}. To detect the car circuit, we run CLA (for $k=5$) on 100 car images sampled uniformly from the ImageNet validation classes: \verb|cab|, \verb|minivan|, \verb|pickup|, and \verb|sports car|. Figure~\ref{fig:teaser} shows the expanded car circuit, which not only recovers all neurons in the original, but surfaces additional units that are also all selective for car features.

\subsection{Intervention analysis of vision models} 
\label{sec:patching}

In order to experimentally measure the effect of circuits on intermediate representations of visual concepts, we define several methods for intervening on circuits in vision models. We can inhibit the identified circuit by \textit{edge pruning}: corrupting (zeroing) all paths between the first and second layers of the circuit (Algorithm \ref{alg:edgeprune}). If the circuit is \textit{exhaustive}, this intervention should cause the model to fail to represent a specific visual concept. To implement this intervention, we (i) run the forward pass, saving ``clean'' activations in the first two model layers containing the circuit (ii) zero the activations of the top layer of neurons in the circuit (iii) run the model in this corrupted form (iv) overwrite the activations of all neurons in the second layer \textit{outside} of the circuit with the clean activations from (i). We complete this modified forward pass from (iv), and study model outputs.  This procedure prevents information from flowing through the first two layers of the circuit, while leaving the rest of the model unaffected. 

To measure the effect of the \textit{entire} circuit on model output, we also define a more aggressive pruning approach, based on edge pruning. We zero activations for all neurons in the \textit{entire} circuit (not just the first two layers) while maintaining clean activations for neurons outside the circuit. In more detail, we (i) run the forward pass, saving ``clean'' activations in \textit{all} model layers containing the circuit  (ii) run a second forward pass, zeroing activations in the circuit neurons, while overwriting all neurons outside the circuit with the activations from (i). We denote this as \textit{circuit pruning} (Algorithm \ref{alg:circuitprune}).

\section{Compositional circuits in Inception-CatFish}
\label{sec:experiments}

Examining circuits also allows us to ask more complex questions regarding intermediate features, such as whether one feature is used to compute another. We study the impact of intermediate circuits (like the car detector from \cite{olah2020zoom} and Section \ref{sec:CLA}) on final model predictions. We construct a dataset where output classes are built from known intermediate concepts, and train a classifier to predict composite classes. CLA enables us to locate groups of neurons representing intermediate concepts that causally affect model behavior. Furthermore, the neurons found with CLA correspond to circuits that are \textbf{re-used} across different output classes.

\vspace{-1mm}
\subsection{Constructing a dataset with visual feature hierarchy (CatFish)}
\label{sec:catfish}

\begin{figure}[t!]
    \centering
    \includegraphics[width=\linewidth]{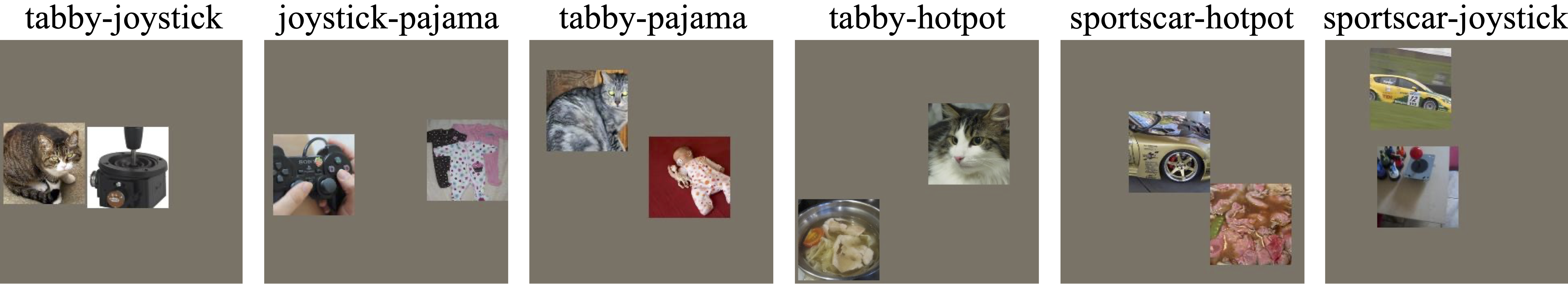}
    \vspace{-5mm}
    \caption{\textbf{CatFish dataset examples.} Each CatFish class composes images from two ImageNet classes. CatFish contains 20 composed classes sampled from 10 ImageNet classes.} 
    \label{fig:catfishex}
    \vspace{-2mm}
\end{figure}

We seek to evaluate whether CLA can recover circuits in image classification models that form class predictions using a ground-truth visual concept hierarchy. To create such models, we construct a dataset (CatFish) by logically composing known concepts. 
The CatFish dataset contains composite images constructed by sampling images from two ImageNet categories and placing them on a neutral background (see Figure \ref{fig:catfishex} for labeled examples and Appendix~\ref{sec:dataset} for more details on dataset construction). A trained model can learn to use the ``intermediate concepts'' for final predictions; CatFish class labels at the output layer (e.g. \verb|tabby-pajama|) can be determined by detecting concepts (e.g. \verb|tabby|, \verb|pajama|) in an intermediate layer. We fine-tune InceptionV1 on CatFish to detect composite CatFish classes, and use CLA to recover circuits corresponding to intermediate concepts inside the trained model (Inception-Catfish). Additional details on model training are provided in Appendix~\ref{sec:inception-catfishtraining}.

\vspace{-2mm}
\subsection{CLA identifies intermediate concept circuits that causally affect model output}
 
We first attempt to probe the representation of visual hierarchy in Inception-CatFish by finding the circuits that detect the lowest-level known features: the intermediate concepts. To find these circuits, we run CLA on 25 input images per concept, generated from pairs of samples of the \textit{same} concept (e.g. \verb|tabby-tabby|). We focus on the final three layers of Inception-CatFish, where Network Dissection \cite{bau2017network} reveals interpretable features.

We also evaluate circuits constructed using three alternative methods: randomly selecting neurons in each layer, selecting the set of neurons in each layer with the largest magnitude \textit{activations} on sample inputs, and those with the largest \textit{weight magnitudes} (following \cite{olah2020zoom}). 

\begin{figure}[t!]
    \centering
    \includegraphics[width=\linewidth]{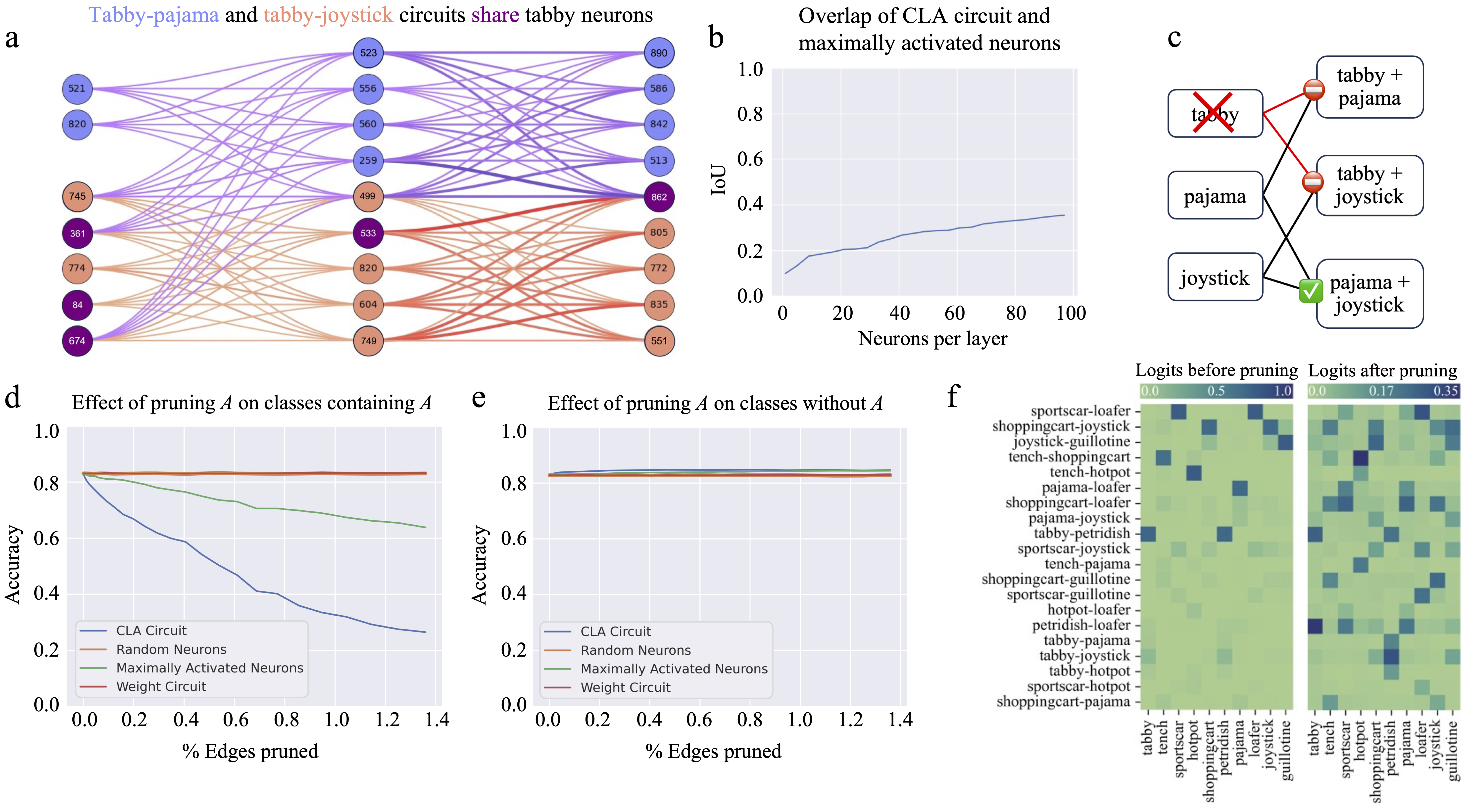}
    \vspace{-6mm}
    \caption{\textbf{Intervening on feature composition in Inception-CatFish by edge pruning.} \textbf{(a)} CLA-generated circuits for the \texttt{tabby-joystick} (red) and \texttt{tabby-pajama} (blue) output classes. Neurons in both circuits (purple) correspond to the shared \texttt{tabby} intermediate concept (see Figure \ref{fig:sharedneurons} for visualization). \textbf{(b)} IoU across neurons selected using CLA, and the maximally activated neurons for a given concept. \textbf{(c)} Predicted impact of intermediate concept knockout. Eliminating a concept through edge pruning will only affect class outputs containing that concept, with no effect on recognition of other concepts. \textbf{(d)} Model accuracy on positive CatFish classes. \textbf{(e)} Model accuracy on negative classes. \textbf{(f)} Inception-CatFish prediction logits before and after pruning each concept circuit, on images sampled from a class containing the concept.}
    \label{fig:catfish_placeholder}
    \vspace{-4mm}
\end{figure}

For each candidate circuit, we measure class prediction accuracy as a proxy for downstream effect size after edge-pruning (Algorithm \ref{alg:edgeprune}) each set of neurons (e.g. the \verb|tabby| circuit for a given $k$, or the $k$ highest-activation units per layer). We separate all 20 CatFish classes into two groups, a ``positive'' set which contains the concept (e.g. \verb|tabby-pajama|, \verb|tabby-petridish|, \verb|tabby-joystick| etc.), and a ``negative'' set of all other classes (e.g. \verb|shoppingcart-pajama|, \verb|tench-hotpot| etc). Figure \ref{fig:catfish_placeholder}c illustrates the hypothesized effect of pruning on positive and negative CatFish classes. If a given circuit has causal impact on model prediction, we expect the corresponding intermediate concept to be ablated from model predictions when edge pruning is applied. We study a variety of circuit sizes, denoted by a $k$ parameter corresponding to the number of neurons per layer. 

As seen in Figure \ref{fig:catfish_placeholder}d,  pruning both random neurons and circuits selected by weight magnitude has negligible effect on classification performance. Pruning maximally activated neurons, and neurons chosen using CLA both have a significant effect on output predictions, with CLA having much greater effect. From Figure \ref{fig:catfish_placeholder}e, we observe small increases in accuracy on negative classes for both CLA and maximally activated neurons, corresponding to the suppression of incorrect predictions in the positive classes. 

We additionally compare the sets of neurons between CLA and maximally activated neurons with Intersection-over-Union (IoU), plotted in Figure \ref{fig:catfish_placeholder}b. We see high divergence across these neurons, indicating the presence of neurons with large activations on specific inputs, but low effect on model outputs. Using the gradient term of our calculated attribution scores, we directly select against such neurons, more accurately recovering the circuit.

\subsection{Pruning an intermediate concept circuit removes that concept from the output distribution}

How do we verify that we are locating and removing a particular intermediate concept, without any effects on the rest of the model? What does Inception-CatFish ``see'' when run on an image from a class containing an ablated concept? To answer such questions, we directly inspect post-ablation model logits across output classes. Specifically, for each intermediate concept, we measure model logits before and after performing edge pruning in Figure \ref{fig:catfish_placeholder}f. After performing edge pruning of an intermediate concept circuit, the probability mass is split; the model allocates roughly equal probability to all output classes containing the \textit{complement} of the pruned intermediate concept. For example, when we prune the \verb|sportscar| circuit and input \verb|sportscar-loafer| images, Inception-CatFish is roughly equally likely to predict any \verb|loafer|-containing class (e.g. \verb|sportscar-loafer, pajama-loafer, shoppingcart-loafer|). Thus, we illustrate that CLA recovers intermediate concept circuits, which are \textit{shared} across output classes containing that concept.

\subsection{Circuits corresponding to CatFish output classes contain intermediate concept neurons}

What is the mechanism by which intermediate concepts impact model predictions? One hypothesis is that the compositional relations (e.g. \verb|tabby| + \verb|pajama| =\verb|tabby-pajama|) are implemented through the direct composition of circuits, i.e. combining their respective neuron sets. To test this, we construct circuits corresponding to composite CatFish output classes (e.g. \verb|tabby-pajama|), by running CLA on 25 input images drawn from the output class (see Figure \ref{fig:catfish_placeholder}a, figure \ref{fig:sharedneurons} for visualizations). Using IoU \cite{jaccardIOU} to compare neuron sets, we find that the CatFish class circuits from CLA are \textit{built} from individual concept circuits (Figure \ref{fig:iou-catfish}.) The high overlap between the concept circuits and their corresponding CatFish class circuits (e.g. [\verb|tabby| neurons $\cup$ \verb|pajama| neurons] $\cap$ \verb|tabby-pajama| neurons ) suggests that intermediate concept circuits \textit{directly compose} to form CatFish class circuits, which are responsible for output behavior.

\section{Circuit pruning defends CLIP from text-based adversarial attacks}
\label{sec:clip}
A potential use case of model editing is the defense of models from spurious correlations and unwanted features in data. Previous work has found that large multimodal models such as CLIP are vulnerable to adversarial attacks from natural images containing text conflicting with image content \cite{goh2021multimodal}. For example, an apple with a paper saying ``iPod'' taped onto it might be misclassified as an iPod. To prevent such adversarial attacks, we propose performing interventions based on the underlying decision-making pipeline of the model, as discovered by circuit analysis. Previous approaches to defending such attacks include MILAN \cite{hernandez2022natural}, wherein all units that appear to recognize text in ImageNet validation images are ablated. Our more efficient method only requires inference on a small sample of 50 images and performs small interventions that are minimally destructive to model performance. 


\subsection{Traffic light dataset for benchmarking textual defense}
\begin{figure}[h]
    \centering
    \includegraphics[width=0.8\linewidth]{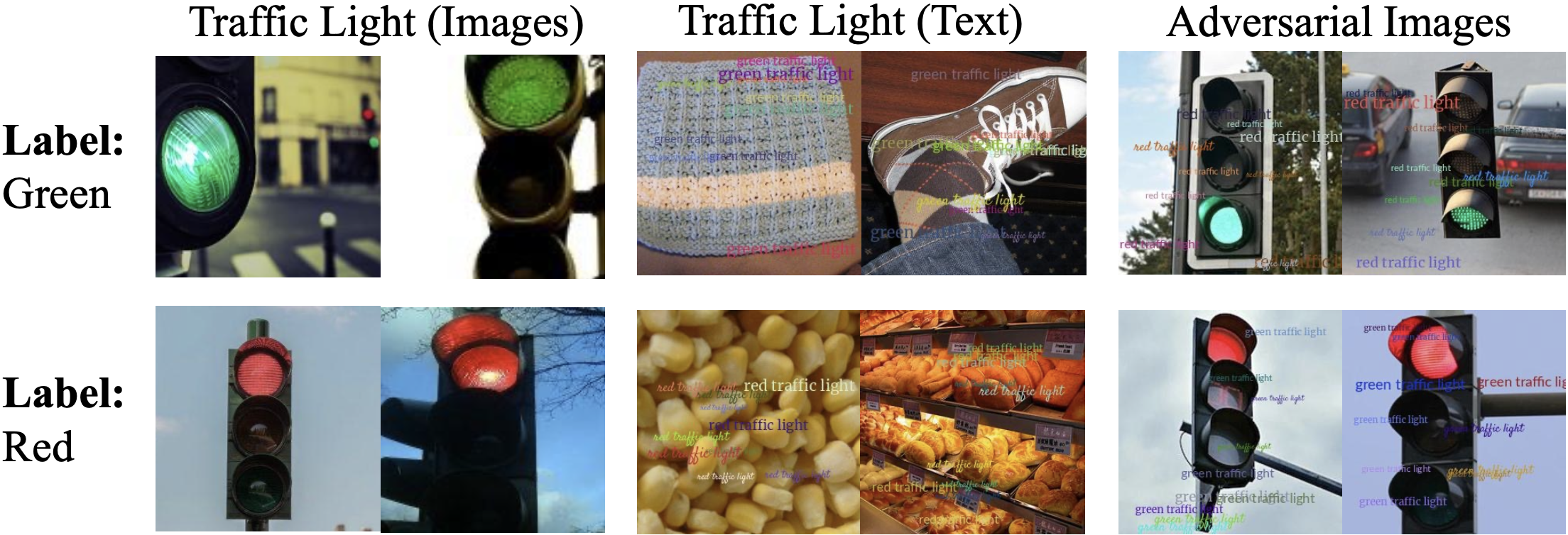}
    \caption{\textbf{Samples from the Traffic Light dataset.} The dataset includes real traffic light images, ImageNet with overlaid traffic light text, and adversarial text-attacked images.}
    \label{fig:traffic_light}
\end{figure}

To benchmark the performance of circuit interventions, we propose an example scenario: the case of using CLIP to label traffic lights based on their color (red or green), in the presence of potential text-based adversarial attacks. We construct a dataset of 50 training images for circuit identification and 100 testing images for circuit validation for three classes of images: (i) red/green traffic lights (found using CLIP retrieval of ``red traffic light'' or ``green traffic light'' on LAION-5B) (ii) ImageNet validation images with instances of ``red traffic light'' or ``green traffic light'' overlaid with random font size, color, and position, and (iii) holdout adversarial images with red/green traffic lights with overlaid text indicating the opposite color. Figure~\ref{fig:traffic_light} shows several example images from the dataset.

\vspace{-3mm}
\subsection{Model intervention protects CLIP from adversarial attacks}
\begin{figure}[h]
    \centering
    \includegraphics[width=.95\linewidth]{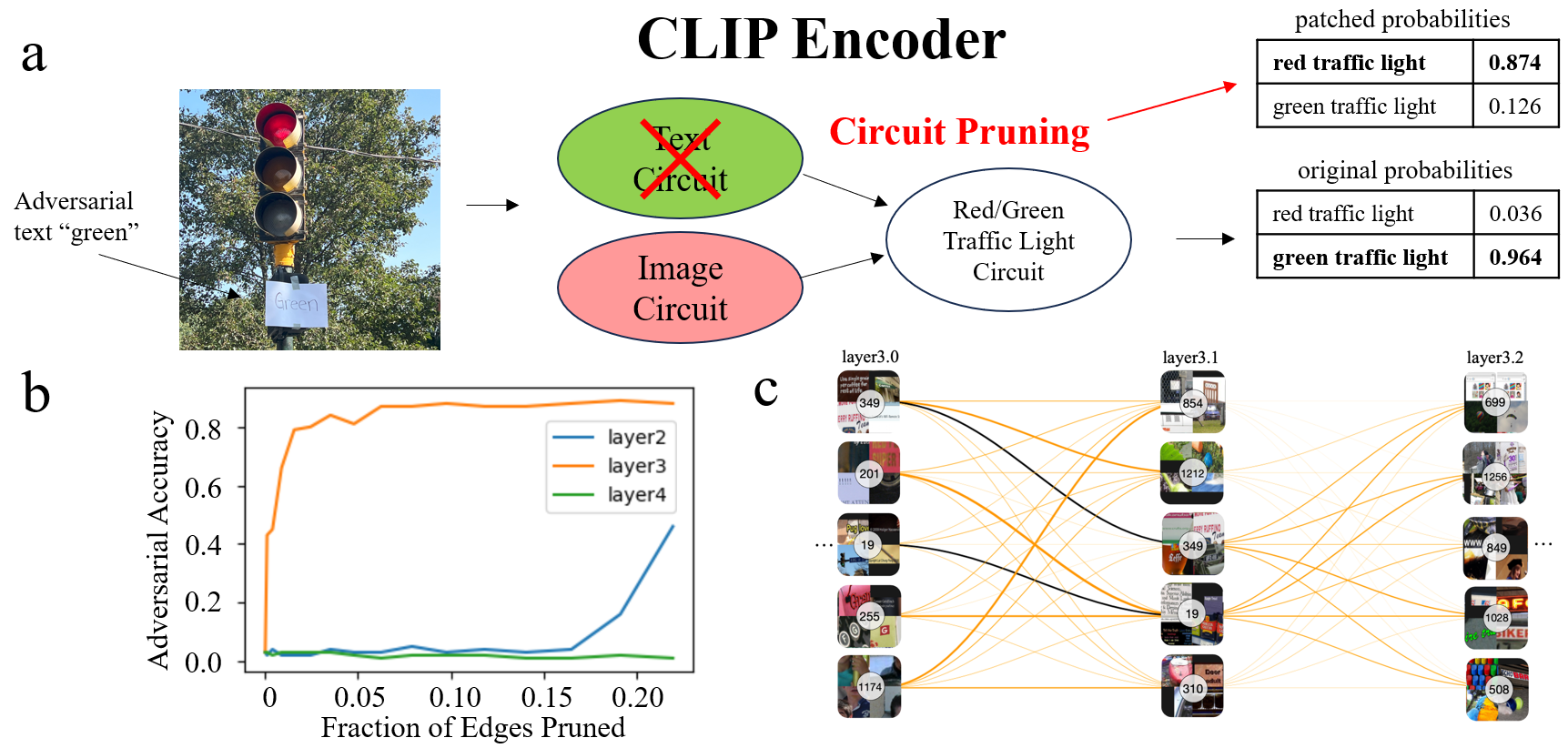}
    \caption{\textbf{Intervening on CLIP to prevent text-based adversarial attacks. (a)} Schematic of the intervention showing that pruning the text circuit defends CLIP from a real-world adversarial attack.  \textbf{(b)} Adversarial accuracy as a function of circuit width when intervening on three CLIP layers \textbf{(c)} CLA text circuit for layer 3. Residual connections between equivalent channels are shown in black.}
    \label{fig:clip}
\end{figure}

Fundamentally, the main ``traffic light'' classification circuit in CLIP is composed of at least two subcircuits: an image detector that classifies real traffic lights, and a text detector that detects and classifies text. We find the text detector automatically using CLA, and then use circuit pruning (Algorithm~\ref{alg:circuitprune}) to remove the text detector from the model. We perform a sweep of CLIP layer (2, 3, or 4) and the width of the text circuit ($k$); results are shown in Figure~\ref{fig:clip}b. On the full test set, the accuracy of CLIP on adversarial images improves from $3\%$ to $87\%$ while pruning only $6\%$ of edges in layer 3. Thus, the intervention successfully defends against text-based adversarial attacks. Examples of neurons for each circuit are shown in Figure~\ref{fig:clip}c.

\vspace{-3mm}
\section{Discussion}

We introduce a method for automatically discovering circuits in vision networks. We perform experiments on Inception-CatFish, a model with induced visual feature hierarchy, and CLIP, a multimodal foundation model. CLA identifies subgraphs for computation over intermediate concepts. The primary limitation of CLA is its strict topological restriction on circuit shape. Future work should allow for automatic selection of different numbers of neurons per layer.

\vspace{-3mm}
\section{Acknowledgements}

We are grateful for the support of the MIT-IBM Watson AI Lab, and ARL grant W911NF-18-2-0218. We thank David Bau, Tamar Rott Shaham, and Tazo Chowdhury for their useful input and insightful discussions.


\bibliographystyle{plain}
\bibliography{refs}  

\clearpage
\appendix

\renewcommand{\thetable}{A\arabic{table}} 
\setcounter{subsection}{0}
\renewcommand{\thesubsection}{A\arabic{subsection}}
\renewcommand{\thefigure}{A\arabic{figure}} 
\setcounter{section}{0}
\setcounter{subsection}{0}
\renewcommand{\thesubsection}{\Alph{section}\arabic{subsection}} 

\textbf{\Large {Appendix}}

\section{Automatic circuit discovery in pretrained InceptionV1}
\label{sec:pretrainedinception}
We perform most of our experiments on an Inception model trained on a dataset with a ground-truth visual hierarchy. We also are interested in whether CLA can detect circuits in a generic InceptionV1 model \cite{szegedy2015going} trained on ImageNet. We detected circuits using 100 input images sampled across four car-containing ImageNet classes (25 each): \verb|cab|, \verb|minivan|, \verb|pickup|, and \verb|sports car|, using layers 4b, 4c, and 4d. To test the effect of ablation, we perform path patching on every path in the circuit as outlined in \cite{pathpatch}, replacing input activations with those from a randomly chosen image (drawn from the \verb|banana| class). We use KL divergence~\cite{kldiv} as a measure of effect on model output, taken against the ground truth outputs, over a set of images drawn from the \verb|sports car| ImageNet class. We compare CLA to other approaches, including using circuits derived from weight magnitudes, output attribution \cite{buildingblocks} to select neurons relevant to the car-containing output classes, and randomly chosen neurons. Figure A1 shows that CLA has a greater effect on model output compared to other methods, corresponding to increased concept erasure.

\begin{figure}[htbp!]
    \centering
    \includegraphics[width=\linewidth]{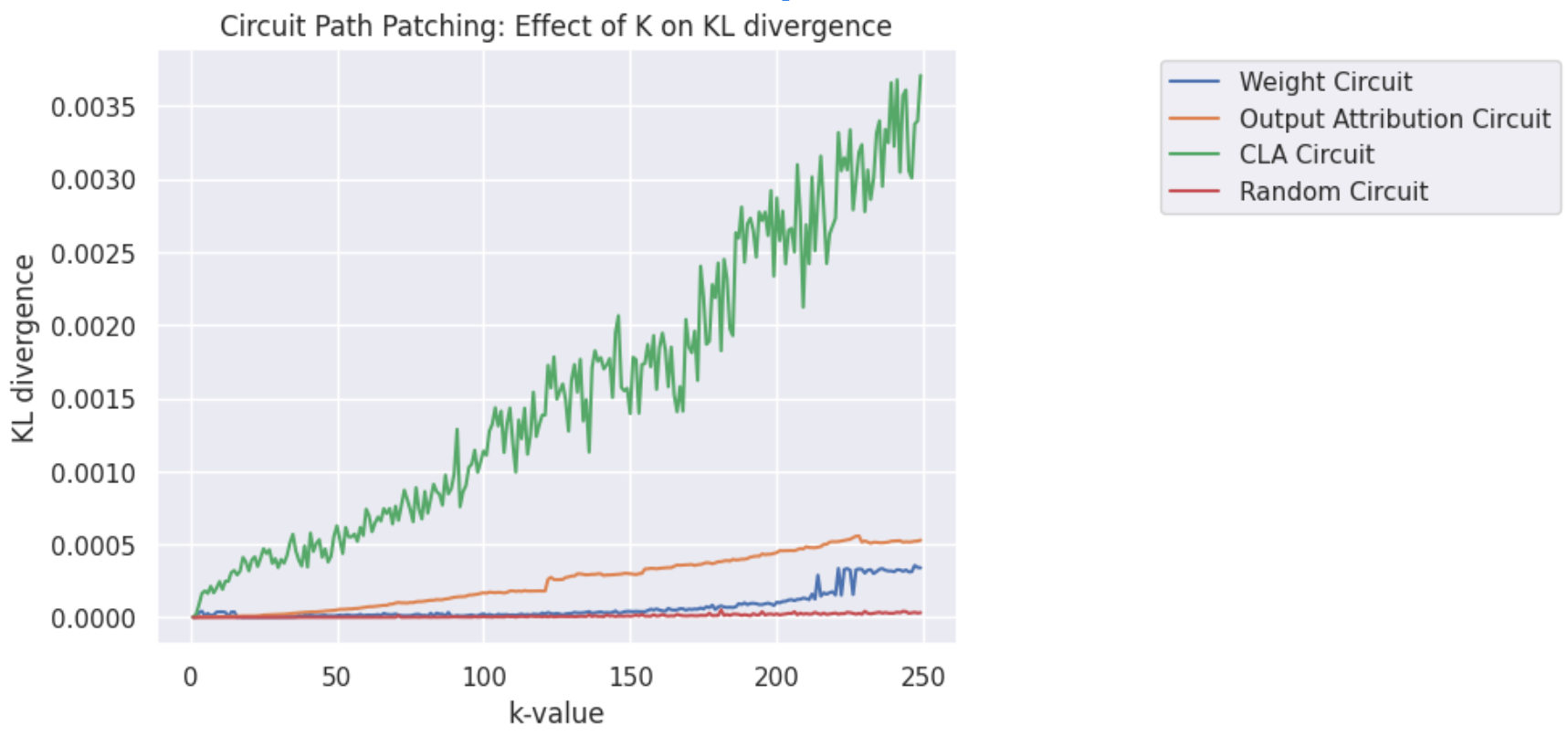}
    \vspace{-5mm}
    \caption{\textbf{CLA circuit maximally affects model performance.} CLA discovers a unified ``car circuit,'' with large effects on model output on  images drawn from the \textit{sports-car} class. The general "car circuit" was found by running CLA on 100 input images sampled across four car-containing ImageNet classes (25 each)} 
    \label{fig:imagenetperf}
\end{figure}

\section{Inception-CatFish training details} 
\label{sec:inception-catfishtraining}
To create Inception-CatFish, we finetune the InceptionV1 model in PyTorch (originally pretrained on ImageNet classification) to classify images in the CatFish dataset with cross-entropy loss. Finetuning was performed using data parallelism across 6 RTX 3090s \cite{torch}. We performed a hyperparameter search over choices of learning rate, batch size, and LR decay. We optimize using SGD with learning rate 0.01 and a momentum of 0.9 \cite{momentum}, and use a batch size of 512. We add a fixed learning rate schedule, specifically a LR decay of 0.5\% per epoch, and train until model convergence. We place Imagenet images directly on a neutral background (Imagenet mean). We then apply standard Imagenet normalization to create the tensors for model training. 

\section{CatFish dataset}
\label{sec:dataset}
\begin{figure}[t!]
    \centering
    \includegraphics[width=\linewidth]{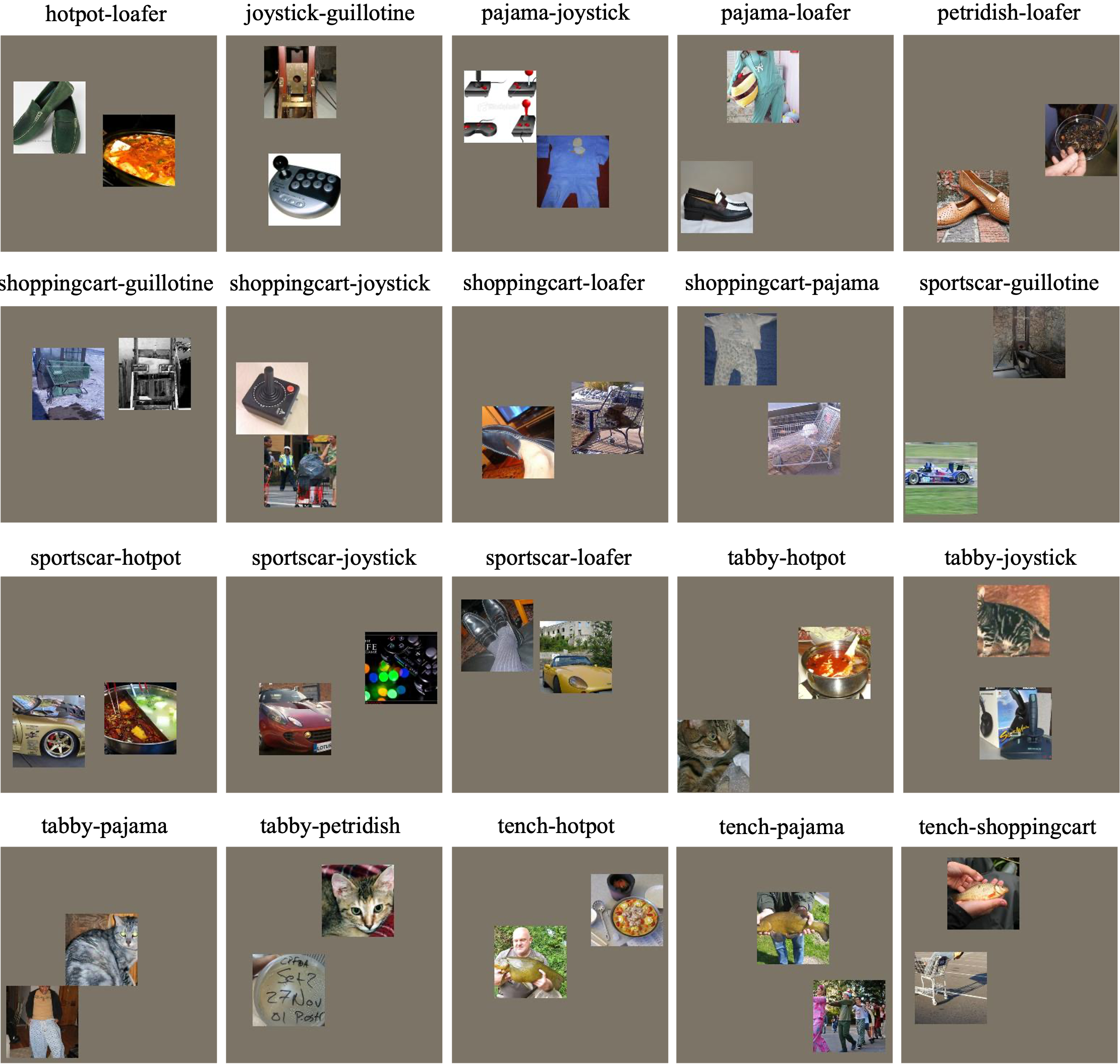}
    \vspace{-5mm}
    \caption{\textbf{Catfish dataset.} Each Catfish class composes images from two ImageNet classes. Catfish contains 20 composed classes sampled from 10 ImageNet classes.} 
    \label{fig:catfish}
\end{figure}

We create 20 pairs from 10 classes hand-selected in order to ensure semantic differences. From each pair of classes, we generate images using the following process: (i) select an image from each class. (ii) rescale each image into 100x100 ``patches.'' (iii) place both patches on a 300x300 background (with the mean color of ImageNet images) ensuring there there is no overlap. Figure~\ref{fig:catfish} shows example images drawn from the CatFish dataset. For each of 20 CatFish classes, we generate 3000 images, for a total of 60,000 training images. For validation and testing, we generate 150 unique images per class, for 3000 validation and test set images. To prevent data leakage, we ensure that these datasets use different sets of source images. 

\section{CLA circuits are stable across values of $k$}
\label{sec:stability}
\begin{figure}[htbp!]
    \centering
    \includegraphics[width=\linewidth]{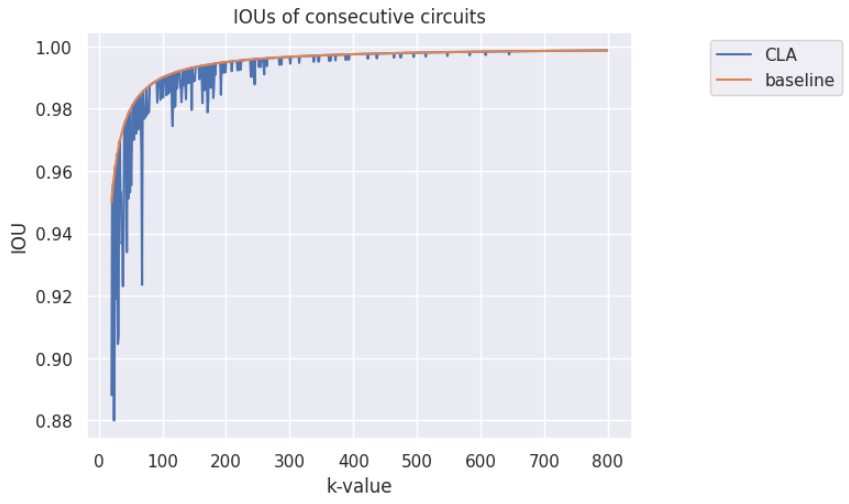}
    \vspace{-5mm}
    \caption{\textbf{Consecutive circuit overlap.} CLA derives circuits that contain similar neurons, independent of the choice of $k$. The baseline indicates the IoU expected if the smaller circuit were entirely contained in the larger circuit.} 
    \label{fig:stable}
\end{figure}

Throughout this work, we generate circuits using different choices of $k$, a parameter denoting the number of neurons per layer in the circuit. In fact, we recompute the circuit from scratch for different $k$ values, meaning that, for example, the neurons included in the circuit for $k=50$ and $k=55$ could vary considerably. To what extent is this the case? We evaluate the extent to which circuits found using increasing values of $k$ build upon a consistent set of neurons by computing the IoU between the sets of neurons derived from consecutive circuits across all sizes. Figure \ref{fig:stable} shows that circuits derived from CLA are ``stable'', and do not depend much on the value of $k$, except for very low values. In all cases, IoU of consecutive circuits never drops below 0.85, indicating very high overlap.

\begin{figure}[h]
     \centering
     \includegraphics[width=0.6\linewidth]{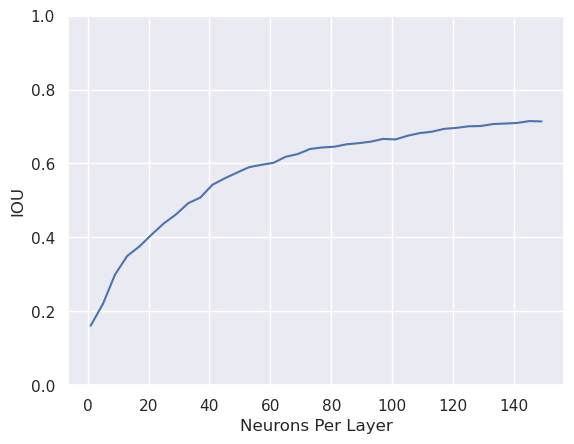}
     \caption{\textbf{Overlap between output class and intermediate subclass circuits:} Aggregated across all classes, the IoU shows high overlap between a class circuit (e.g. \texttt{tabby-pajama}) with $2k$ neurons per layer, and the union of the corresponding two subclass circuits (e.g. \texttt{tabby} $\cup$ \texttt{pajama}), each with $k$ neurons per layer.}
     \label{fig:iou-catfish}
 \end{figure}

\newpage
\section{Gradual removal of concepts by pruning circuits}
\label{sec:partialremoval}
\begin{figure}[b!]
    \centering
    \includegraphics[width=\linewidth]{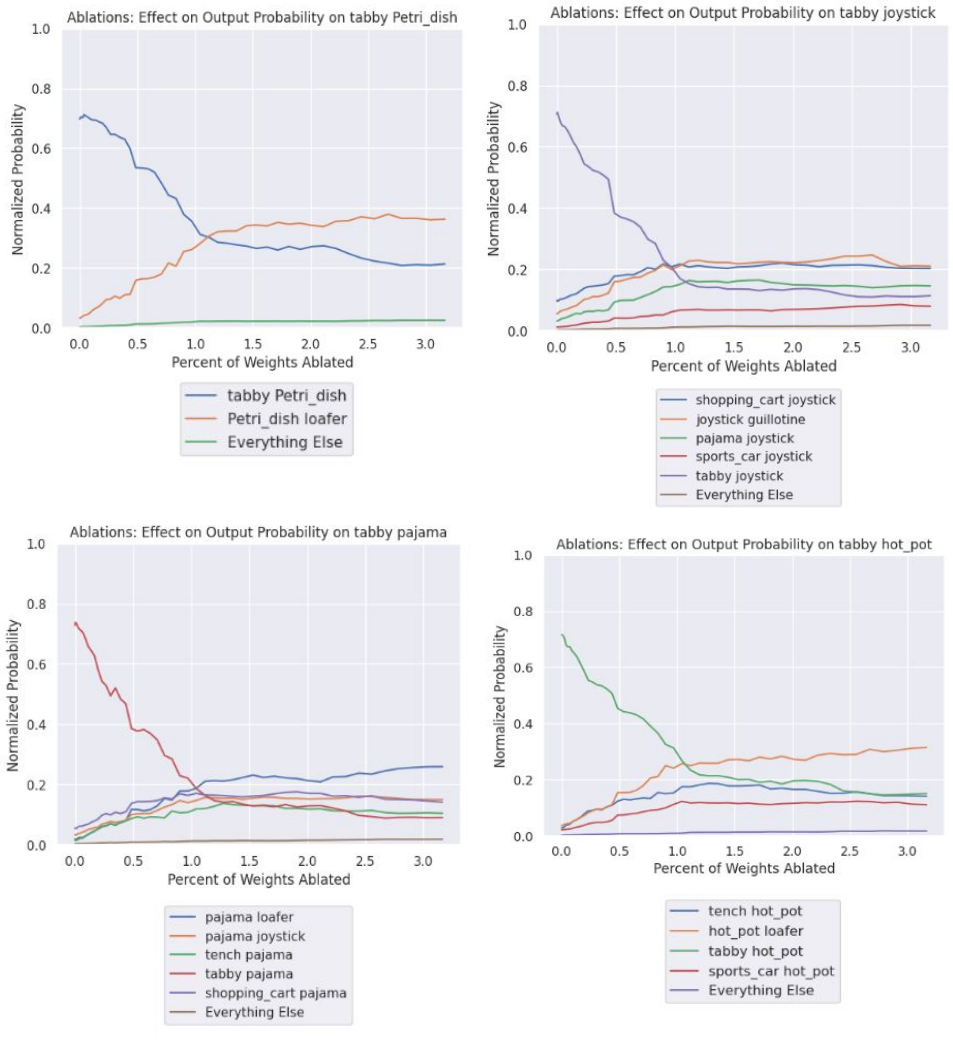}
    \vspace{-5mm}
    \caption{\textbf{ Partial concept ablation.} Varying circuit size, we show that partial intermediate concept ablation is achievable. Specifically, as the size of the ablated ``tabby'' circuit increases, the model "forgets" the correct output class on inputs containing ``tabby'', shown by the gradually decreasing output probability.} 
    \label{fig:partial}
\end{figure}

We explore how removing a circuit corresponding to a CatFish intermediate concept affects the prediction logits. Specifically, we prune circuits for the \verb|tabby| intermediate concept at various sizes and inspect model output logits on all CatFish classes containing \verb|tabby|. We find that as the size of the circuit removed increases, intermediate concepts are \textbf{gradually} ablated, with the output probability for the correct output decreasing. Results of this as a function of circuit size (neurons per layer, indicated as $k$) is shown in Figure \ref{fig:partial}. 

\begin{figure}[b!]
    \centering
    \includegraphics[width=.7\linewidth]{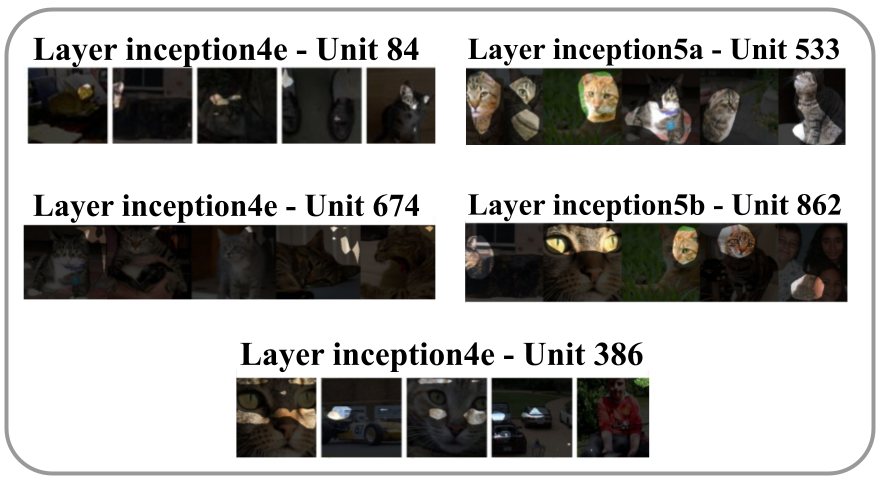}
    \vspace{-2mm}
    \caption{\textbf{Shared tabby neurons.} Computing the overlap between the \texttt{tabby-pajama} and \texttt{tabby-joystick} circuits yields five neurons across three layers. The top five dataset exemplars that cause the greatest activation for each neuron correspond to the shared \texttt{tabby} concept.}
    \label{fig:sharedneurons}
\end{figure}

\begin{algorithm}
\caption{\textbf{Edge Pruning} inhibits a circuit by pruning the edge connections between the first and second layers of the circuit. This prevents information flow through the circuit.}\label{alg:edgeprune}
\begin{algorithmic}[1]

\Procedure{EdgePrune}{model, input, circuit, layers $\{l_1, l_2\}$}
    \State clean\_activations $\leftarrow$ model(image) \Comment{Get clean activations from model}

    \Return model(input) with intervention for all $l$:
     \State activations[$l_1$] = model.$l_1$(input)
     \For{each channel $c$ in layer $l_1$}
        \If{channel $c$ $\in$ circuit}
            \State activations[$l_1, c$] = 0 \Comment{Ablate neurons in circuit}
        \EndIf
    \EndFor

    \State activations[$l_2$] = model.$l_2$(activations)
     \For{each channel $c$ in layer $l_2$}
        \If{channel $c$ $\not\in$ circuit}
            \State activations[$l_2, c$] = clean\_activations[$l_2, c$] \Comment{Preserve neurons not in circuit}
        \EndIf
    \EndFor

    \Return model.output(activations) \Comment{Proceed from layer $l_2$}
\EndProcedure

\end{algorithmic}
\end{algorithm}

\begin{algorithm}
\caption{\textbf{Circuit Pruning} removes all neurons in a circuit, cutting it off from the rest of the model. This is achieved by repeating the edge ablation process across all layers, and all edges.}\label{alg:circuitprune}
\begin{algorithmic}[1]

\Procedure{CircuitPrune}{model, input, circuit, layers $\{l_i\}$}
    \State clean\_activations $\leftarrow$ model(image) \Comment{Get clean activations from model}

    \Return model(input) with intervention for all $l$:
    \For{layer $l_i$ in $\{l_i\}$}
         \State activations = model.$l_i$(activations[$l_{i-1}$])
         \For{each channel $c$ in layer $l_i$}
            \If{channel $c$ $\in$ circuit}
                \State activations[$l_i, c$] = 0 \Comment{Ablate neurons in circuit}
            \ElsIf {channel $c$ $\notin$ circuit}
                \State activations[$l_i$, $c$] = clean\_activations[$l_i$, $c$] \Comment{Preserve neurons outside circuit}
            \EndIf
        \EndFor
        \State model.$l_i$ = activations \Comment{Set newly patched activations}
    \EndFor
\EndProcedure

\end{algorithmic}
\end{algorithm}

\end{document}